\documentclass[runningheads, envcountsame, a4paper]{llncs}

\usepackage{subfigure}
\usepackage[pdftex]{graphicx}
\usepackage{epstopdf}
\usepackage{times}
\usepackage{graphicx}
\usepackage{amsmath}
\usepackage{amssymb}
\usepackage{multirow}
\usepackage{svg}
\usepackage[hyphens]{url}
\usepackage[misc]{ifsym}
\usepackage{booktabs}       
\usepackage{microtype}
\newcommand{\tabincell}[2]{\begin{tabular}{@{}#1@{}}#2\end{tabular}}  
\def\bs{\boldsymbol}

\begin{document}
\title{$L_0$-ARM: Network Sparsification via Stochastic Binary Optimization}
\titlerunning{$L_0$-ARM: Network Sparsification via Stochastic Binary Optimization}
%
\author{Yang Li \and
Shihao Ji~\Letter}
\authorrunning{Y. Li and S. Ji}
%
\institute{Georgia State University, \, USA\\
\email{yli93@student.gsu.edu, sji@gsu.edu}}
\tocauthor{Yang~Li and Shihao~Ji}
\toctitle{$L_0$-ARM: Network Sparsification via Stochastic Binary Optimization}
\maketitle              

\begin{abstract}
  We consider network sparsification as an $L_0$-norm regularized binary optimization problem, where each unit of a neural network (e.g., weight, neuron, or channel, etc.) is attached with a stochastic binary gate, whose parameters are jointly optimized with original network parameters. The Augment-Reinforce-Merge (ARM)~\cite{Yin2019}, a recently proposed  unbiased gradient estimator, is investigated for this binary optimization problem. Compared to the hard concrete gradient estimator from Louizos et al. ~\cite{Louizos2017}, ARM demonstrates superior performance of pruning network architectures while retaining almost the same accuracies of baseline methods. Similar to the hard concrete estimator, ARM also enables conditional computation during model training but with improved effectiveness due to the \emph{exact} binary stochasticity. Thanks to the flexibility of ARM, many smooth or non-smooth parametric functions, such as scaled sigmoid or hard sigmoid, can be used to parameterize this binary optimization problem and the unbiasness of the ARM estimator is retained, while the hard concrete estimator has to rely on the hard sigmoid function to achieve conditional computation and thus accelerated training. Extensive experiments on multiple public datasets demonstrate state-of-the-art pruning rates with almost the same accuracies of baseline methods. The resulting algorithm $L_0$-ARM sparsifies the Wide-ResNet models on CIFAR-10 and CIFAR-100 while the hard concrete estimator cannot. The code is public available at \url{https://github.com/leo-yangli/l0-arm}.
\end{abstract}

\section{Introduction}

Deep Neural Networks (DNNs) have achieved great success in a broad range of applications in image recognition~\cite{imagenet09}, natural language processing~\cite{bert18}, and games~\cite{alphago16}. Latest DNN architectures, such as ResNet~\cite{resnet16}, DenseNet~\cite{densenet17} and Wide-ResNet~\cite{zagoruyko2016wide}, incorporate hundreds of millions of parameters to achieve state-of-the-art predictive performance. However, the expanding number of parameters not only increases the risk of overfitting, but also leads to high computational costs. Many practical real-time applications of DNNs, such as for smart phones, drones and the IoT (Internet of Things) devices, call for compute and memory efficient models as these devices typically have very limited computation and memory capacities. 

Fortunately, it has been shown that DNNs can be pruned or sparsified significantly with minor accuracy losses~\cite{han2015learning,han2015deep}, and sometimes sparsified networks can even achieve higher accuracies due to the regularization effects of the network sparsification algorithms~\cite{NekMolAsh17,Louizos2017}. Driven by the widely spread applications of DNNs in real-time systems, there has been an increasing interest in pruning or sparsifying networks recently~\cite{han2015learning,han2015deep},  \cite{WenWuWan16,li2016pruning,louizos2017bayesian,molchanov2017variational,NekMolAsh17,Louizos2017}. Earlier methods such as the magnitude-based approaches~\cite{han2015learning,han2015deep} prune networks by removing the weights of small magnitudes, and it has been shown that this approach although simple is very effective at sparsifying network architectures with minor accuracy losses. Recently, the $L_0$-norm based regularization method~\cite{Louizos2017} is getting attraction as this approach \textit{explicitly} penalizes number of non-zero parameters and can drive redundant or insignificant parameters to be exact zero. However, the gradient of the $L_0$ regularized objective function is intractable. Louizos et al.  ~\cite{Louizos2017} propose to use the hard concrete distribution as a close surrogate to the Bernoulli distribution, and this leads to a differentiable objective function while still being able to zeroing out redundant or insignificant weights during training. Due to the hard concrete substitution, however, the resulting hard concrete estimator is biased with respect to the original objective function. 

In this paper, we propose $L_0$-ARM for network sparsification. $L_0$-ARM is built on top of the $L_0$ regularization framework of Louizos et al.  ~\cite{Louizos2017}. However, instead of using a biased hard concrete gradient estimator, we investigate the Augment-Reinforce-Merge (ARM)~\cite{Yin2019}, a recently proposed unbiased gradient estimator for stochastic binary optimization. Because of the unbiasness and flexibility of the ARM estimator, $L_0$-ARM exhibits a significantly faster rate at pruning network architectures and reducing FLOPs than the hard concrete estimator. Extensive experiments on multiple public datasets demonstrate the superior performance of $L_0$-ARM at sparsifying networks with fully connected layers and convolutional layers. It achieves state-of-the-art prune rates while retaining similar accuracies compared to baseline methods. Additionally, it sparsifies the Wide-ResNet models on CIFAR-10 and CIFAR-100 while the original hard concrete estimator cannot. 

The remainder of the paper is organized as follows. In Sec.~\ref{sec:formulation} we describe the $L_0$ regularized empirical risk minimization for network sparsification and formulate it as a stochastic binary optimization problem. A new unbiased estimator to this problem $L_0$-ARM is presented in Sec.~\ref{sec:algo}, followed by related work in Sec.~\ref{sec:related}. Example results on multiple public datasets are presented in Sec.~\ref{sec:exp}, with comparisons to baseline methods and the state-of-the-art sparsification algorithms. Conclusions and future work are discussed in Sec.~\ref{sec:conclusion}.

\section{Formulation}\label{sec:formulation}

Given a training set $D = \left\{ \left( \bs { x } _ { i } , y_i \right) , i = 1,2 , \cdots , N \right\}$, where $\bs { x_i }$ denotes the input and $y_i$ denotes the target, a neural network is a function $h(\bs x; \bs { \theta })$ parametrized by $\bs { \theta }$ that fits to the training data $D$ with the goal of achieving good generalization to unseen test data. To optimize $\bs { \theta }$, typically a regularized empirical risk is minimized, which contains two terms -- a data loss over training data and a regularization loss over model parameters. Empirically, the regularization term can be weight decay or Lasso, i.e., the $L_2$ or $L_1$ norm of model parameters. 

Since the $L_2$ or $L_1$ norm only imposes shrinkage for large values of $\bs{\theta}$, the resulting model parameters $\bs{\theta}$ are often manifested by smaller magnitudes but none of them are exact zero. Intuitively, a more appealing alternative is the $L_0$ regularization since the $L_0$-norm measures \textit{explicitly} the number of non-zero elements, and minimizing of it over model parameters will drive the redundant or insignificant weights to be exact zero. With the $L_0$ regularization, the empirical risk objective can be written as
  \begin{equation}\label{eq:risk}
      \mathcal { R } ( \bs { \theta } ) = \frac { 1 } { N } \sum _ { i = 1 } ^ { N } \mathcal { L } \left( h ( \bs { x } _ { i } ; \bs { \theta } ) , y_i \right) + \lambda \| \bs { \theta } \| _ { 0 }
  \end{equation} 
where $\mathcal { L } (\cdot)$ denotes the data loss over training data $D$, such as the cross-entropy loss for classification or the mean squared error (MSE) for regression, and $\| \bs { \theta } \| _ { 0 } $ denotes the $L_0$-norm over model parameters, i.e., the number of non-zero weights, and $\lambda$ is a regularization hyper-parameter that balances between data loss and model complexity. 

To represent a sparsified model, we attach a binary random variable $z$ to each element of model parameters $\bs{\theta}$. Therefore, we can re-parameterize the model parameters $\bs{\theta}$ as an element-wise product of non-zero parameters $\tilde{\bs{\theta}}$ and binary random variables $\bs{z}$:
  \begin{equation}\label{eq:theta}    
      \bs { \theta }  = \bs { \tilde { \theta } }  \odot \bs { z },
  \end{equation} 
where $ \bs z  \in \{ 0,1 \} ^ { | \bs { \theta } | } $, and $\odot$ denotes the element-wise product. As a result, Eq.~\ref{eq:risk} can be rewritten as:
\begin{align}\label{eq:R}
      \mathcal { R } ( \tilde { \bs { \theta } }, \bs{z} ) = \frac { 1 } { N } \sum _ { i = 1 } ^ { N } \mathcal { L } \left( h \bigl( \bs { x } _ { i } ; \tilde { \bs { \theta } } \odot \bs { z } \bigr) , y_i \right) + \lambda \sum _ { j = 1 } ^ { | \bs { \tilde { \theta } } | } \bs { 1 }_{\left[  z_j \neq 0 \right]},
\end{align}
where $\bs { 1 }_{[c]}$ is an indicator function that is $1$ if the condition $c$ is satisfied, and $0$ otherwise. Note that both the first term and the second term of Eq.~\ref{eq:R} are not differentiable w.r.t. $\bs { z }$. Therefore, further approximations need to be considered.

According to stochastic variational optimization~\cite{SVO18}, given any function $ \mathcal { F } (\bs{z})  $ and any distribution $q(\bs{z})$, the following inequality holds
\begin{equation}\label{eq:Ele}
      \min_{\bs{z}} \mathcal { F } (\bs{z}) \leq \mathbb{E}_{\bs{z}\sim q(\bs{z})} [\mathcal { F }(\bs{z})],
  \end{equation} 
i.e., the minimum of a function is upper bounded by the expectation of the function. With this result, we can derive an upper bound of Eq.~\ref{eq:R} as follows.

Since $z_j, \forall j\in\{1,\cdots,|\bs{\theta}|\}$ is a binary random variable, we assume $z_j$ is subject to a Bernoulli distribution with parameter $\pi_j\in[0, 1]$, i.e. $z_j\sim \mathrm { Ber } (z;\pi_j)$. Thus, we can upper bound $\min_{\bs{z}} \mathcal { R } ( \tilde { \bs { \theta } }, \bs{z} )$ by the expectation
\begin{align}\label{eq:rhat}
\mathcal {\hat{R}} ( \tilde { \bs { \theta } } , \bs { \pi } ) 
&  =\mathbb { E }_ { \bs { z } \sim  \mathrm{ Ber  } ( \bs { z } ; \bs { \pi } ) } \mathcal { R } ( \tilde { \bs { \theta } } , \bs{z}) \nonumber\\ 
      & = \mathbb { E } _ { \bs { z } \sim  \mathrm{ Ber } ( \bs { z } ; \bs { \pi } ) } \left[ \frac { 1 } { N } \sum _ { i = 1 } ^ { N } \mathcal { L } \left( h ( \bs { x } _ { i } ; \tilde { \bs { \theta } } \odot \bs { z } ) , y_i \right)  \right] + \lambda \sum _ { j = 1 } ^ { | \tilde{\bs\theta} | } \pi _ { j }.
\end{align}
As we can see, now the second term is differentiable w.r.t. the new model parameters $\bs {\pi} $, while the first term is still problematic since the expectation over a large number of binary random variables $\bs {z}$ is intractable and so its gradient. Since $\bs {z} $ are binary random variables following a Bernoulli distribution with parameters $\bs {\pi} $, we now formulate the original $L_0$ regularized empirical risk~(\ref{eq:risk}) to a stochastic binary optimization problem~(\ref{eq:rhat}).

Existing gradient estimators for this kind of discrete latent variable models include REINFORCE~\cite{reinforce92}, Gumble-Softmax~\cite{gumbel-softmax17,concrete17}, REBAR~\cite{rebar17}, RELAX~\cite{relax18} and the Hard Concrete estimator~\cite{Louizos2017}. However, these estimators either are biased or suffer from high variance or computationally expensive due to auxiliary modeling. Recently, the Augment-Reinforce-Merge (ARM)~\cite{Yin2019} gradient estimator is proposed for the optimization of binary latent variable models, which is unbiased and exhibits low variance. Extending this gradient estimator to network sparsification, we find that ARM demonstrates superior performance of prunning network architectures while retaining almost the same accuracies of baseline models. More importantly, similar to the hard concrete estimator, ARM also enables conditional computation~\cite{BenLeoCou18} that not only sparsifies model architectures for inference but also accelerates model training. 

\section{$L_0$-ARM: Stochastic Binary Optimization}\label{sec:algo}
To minimize Eq.~\ref{eq:rhat}, we propose $L_0$-ARM, a stochastic binary optimization algorithm based on the Augment-Reinforce-Merge (ARM) gradient estimator~\cite{Yin2019}. We first introduce the main theorem of ARM. Refer readers to~\cite{Yin2019} for the proof and other details.

\begin{theorem} (ARM)~\cite{Yin2019}. For a vector of $V$ binary random variables $\bs { z }=\left(z_{1}, \cdots, z_{V}\right)$, the gradient of 
  \begin{equation}
      \mathcal { E } ( \bs { \phi } ) = \mathbb { E } _ { \bs { z } \sim \prod _ { v = 1 } ^ { V } \operatorname { Ber } \left( z _ { v } ; g \left( \phi _ { v } \right) \right) } [ f ( \bs { z } ) ]
  \end{equation} 
w.r.t. $\bs { \phi} = (\phi_1, \cdots, \phi_V)$, the logits of the Bernoulli distribution parameters, can be expressed as 
\begin{align}\label{eq:arm}
          \nabla _ { \phi } \mathcal { E } ( \phi )\!=\!\mathbb { E } _ { \bs { u } \sim \prod _ { v = 1 } ^ { V }\!\!  \operatorname { Uniform } \left( u _ { v } ; 0,1 \right) }\Big[ \big( f ( \bs { 1 } _ {[ \bs { u } > g ( - \phi ) ] }) - f ( \bs { 1 } _ { [ \bs { u } < g ( \bs { \phi } ) ] } ) \big) ( \bs { u } - 1 / 2 )\Big ],
\end{align}
where $\bs { 1 } _ { [ \bs { u } > g ( - \bs { \phi } ) ] } : = \left( \bs { 1 } _ { \left[ u _ { 1 } > g \left( - \phi _ { 1 } \right) \right] } , \cdots , \bs { 1 } _ { \left[ u _ { V } > g \left( - \phi _ { V } \right) \right] } \right) ^ T$ and $g(\phi)=\sigma(\phi)=1/(1+\exp(-\phi))$ is the sigmoid function.   
\end{theorem}

Parameterizing $\pi_j\in[0, 1]$ as $g {(\phi _ { j })}$, Eq.~\ref{eq:rhat} can be rewritten as
\begin{align}\label{eq:L0-obj}
          \mathcal {\hat{R} } ( \tilde { \bs { \theta } } , \bs { \phi } ) &=  \mathbb { E }_ { \bs { z } \sim \mathrm{ Ber } ( \bs { z } ; g (\bs { \phi } ) ) } \left[f(\bs { z })\right] + \lambda \sum _ { j = 1 } ^ { | \tilde{\theta} | } g {(\bs { \phi }  _ { j })}  \nonumber\\
          &=  \mathbb { E }_ { \bs { u } \sim \mathrm { Uniform } ( \bs { u } ; 0, 1) } \left[f(\bs { \bs { 1 } _ { [ \bs { u } < g ( \bs { \phi } ) ] } })\right] + \lambda \sum _ { j = 1 } ^ { | \tilde{\theta} | } g {(\bs { \phi }  _ { j })},
\end{align}
where $f(\bs{z}) = \frac { 1 } { N } \sum _ { i = 1 } ^ { N } \mathcal { L } \left( h ( \bs { x } _ { i } ; \tilde { \bs { \theta } } \odot \bs { \bs{z} }) , y_i \right) $.
Now according to Theorem 1, we can evaluate the gradient of Eq.~\ref{eq:L0-obj} w.r.t. $\bs{\phi}$ by
\begin{align}\label{eq:L0-ARM}
          \nabla^{ARM} _ { \bs { \phi } } \mathcal { \hat{R} } (\tilde { \bs { \theta } }, \bs{\phi} ) &= \mathbb { E } _ { \bs { u } \sim \mathrm { Uniform } \left( \bs{u}; 0,1 \right)} \Big[ \big( f ( \bs { 1 } _ { [ \bs { u } > g ( - \bs{\phi} ) ] } ) - f ( \bs { 1 } _ { [ \bs { u } < g ( \bs { \phi } ) ] }) \big) ( \bs { u } - 1 / 2)\Big] \nonumber\\
          &\quad + \lambda \sum _ { j = 1 } ^ { | \tilde{\theta} | } \nabla_{\phi_j}g {(\phi_{ j })},
\end{align}
which is an unbiased and low variance estimator as demonstrated in~\cite{Yin2019}.

Note from Eq.~\ref{eq:L0-ARM} that we need to evaluate $f(\cdot)$ twice to compute the gradient, the second of which is the same operation required by the data loss of Eq.~\ref{eq:L0-obj}. Therefore, one extra forward pass $f ( \bs { 1 } _ { [ \bs { u } > g ( - \phi ) ] } )$ is required by the $L_0$-ARM gradient estimator. This additional forward pass might be computationally expensive, especially for networks with millions of parameters. To reduce the computational complexity of Eq.~\ref{eq:L0-ARM}, we further consider another gradient estimator -- Augment-Reinforce (AR)~\cite{Yin2019}:
\begin{align}\label{eq:L0-AR}
          \nabla^{AR} _ { \bs { \phi } } \mathcal { \hat{R} } (\tilde { \bs { \theta } }, \bs{\phi} ) &= \mathbb { E } _ { \bs { u } \sim \mathrm { Uniform } \left( \bs{u}; 0,1 \right)} \Big[  f ( \bs { 1 } _ { [ \bs { u } < g ( \bs { \phi } ) ] }) (1- 2\bs { u })\Big] \nonumber\\ &\quad + \lambda \sum _ { j = 1 } ^ { | \tilde{\theta} | } \nabla_{\phi_j}g {(\phi  _ { j })},
\end{align}
which requires only one forward pass  $f ( \bs { 1 } _ { [ \bs { u } < g ( \phi ) ] } )$ that is the same operation as in Eq.~\ref{eq:L0-obj}. This $L_0$-AR gradient estimator is still unbiased but with higher variance. Now with $L_0$-AR, we can trade off the variance of the estimator with the computational complexity. We will evaluate the impact of this trade-off in our experiments.

\subsection{Choice of $g(\phi)$}\label{sec:choice}
\begin{figure}[t] 
\begin{center}
  \includegraphics[width=0.7\linewidth]{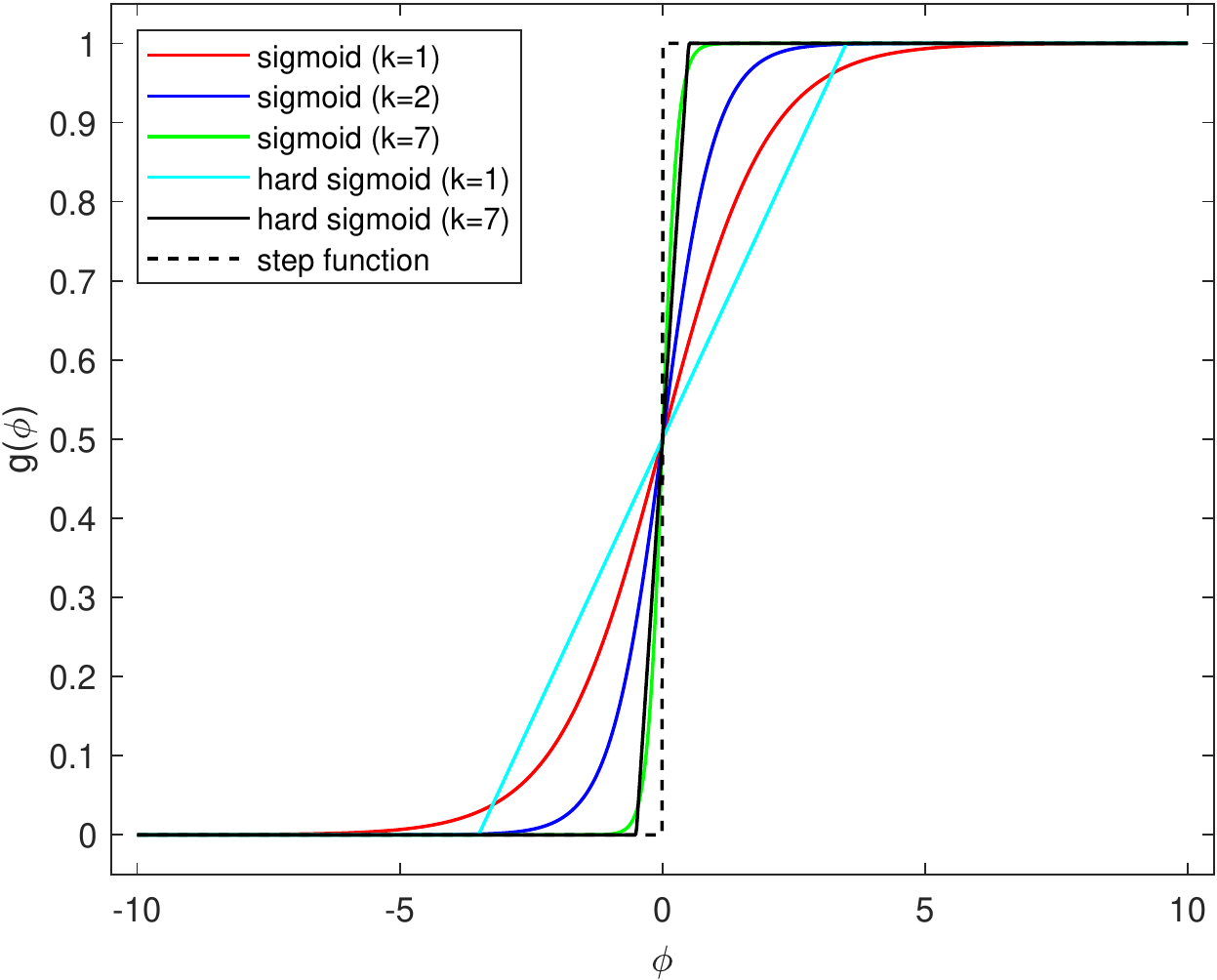}
\end{center}\vspace{-16pt}
\caption{The plots of $g(\phi)$ with different $k$ for sigmoid and hard sigmoid functions. A large $k$ tends to be more effective at sparsifying networks. Best viewed in color.}\label{fig:sigmoid}
\end{figure}

Theorem 1 of ARM defines $g(\phi)=\sigma(\phi)$, where $\sigma(\cdot)$ is the sigmoid function. For the purpose of network sparsification, we find that this parametric function isn't very effective due to its slow transition between values 0 and 1. Thanks to the flexibility of ARM, we have a lot of freedom to design this parametric function $g(\phi)$. Apparently, it's straightforward to generalize Theorem 1 for any parametric functions (smooth or non-smooth) as long as $g: \mathcal{R}\to[0, 1]$ and $g(-\phi)=1-g(\phi)$~\footnote{The second condition is not necessary. But for simplicity, we will impose this condition to select parametric function $g(\phi)$ that is antithetic. Designing $g(\phi)$ without this constraint could be a potential area that is worthy of further investigation.}. Example parametric functions that work well in our experiments are the scaled sigmoid function 
\begin{align}
g_{\sigma_k}(\phi) = \sigma(k\phi)=\frac{1}{1+\exp(-k\phi)},
\end{align}
and the centered-scaled hard sigmoid
\begin{align}
g_{\bar{\sigma}_k}(\phi) = \min(1, \max(0, \frac{k}{7}\phi+0.5)),
\end{align}
where $7$ is introduced such that $g_{\bar{\sigma}_1}(\phi)\approx g_{\sigma_1}(\phi)=\sigma(\phi)$. See Fig.~\ref{fig:sigmoid} for some example plots of $g_{\sigma_k}(\phi)$ and $g_{\bar{\sigma}_k}(\phi)$ with different $k$. Empirically, we find that $k=7$ works well for all of our experiments.

One important difference between the hard concrete estimator from Louizos et al.~\cite{Louizos2017} and $L_0$-ARM is that the hard concrete estimator has to rely on the hard sigmoid gate to zero out some parameters during training (a.k.a. conditional computation~\cite{BenLeoCou18}), while $L_0$-ARM achieves conditional computation naturally by sampling from the Bernoulli distribution, parameterized by $g(\phi)$, where $g(\phi)$ can be any parametric function (smooth or non-smooth) as shown in Fig.~\ref{fig:sigmoid}. We validate this in our experiments.

\subsection{Sparsifying Network Architectures for Inference}\label{sec:inference}
\begin{figure}[t]
\begin{center}
  \includegraphics[width=0.8\linewidth]{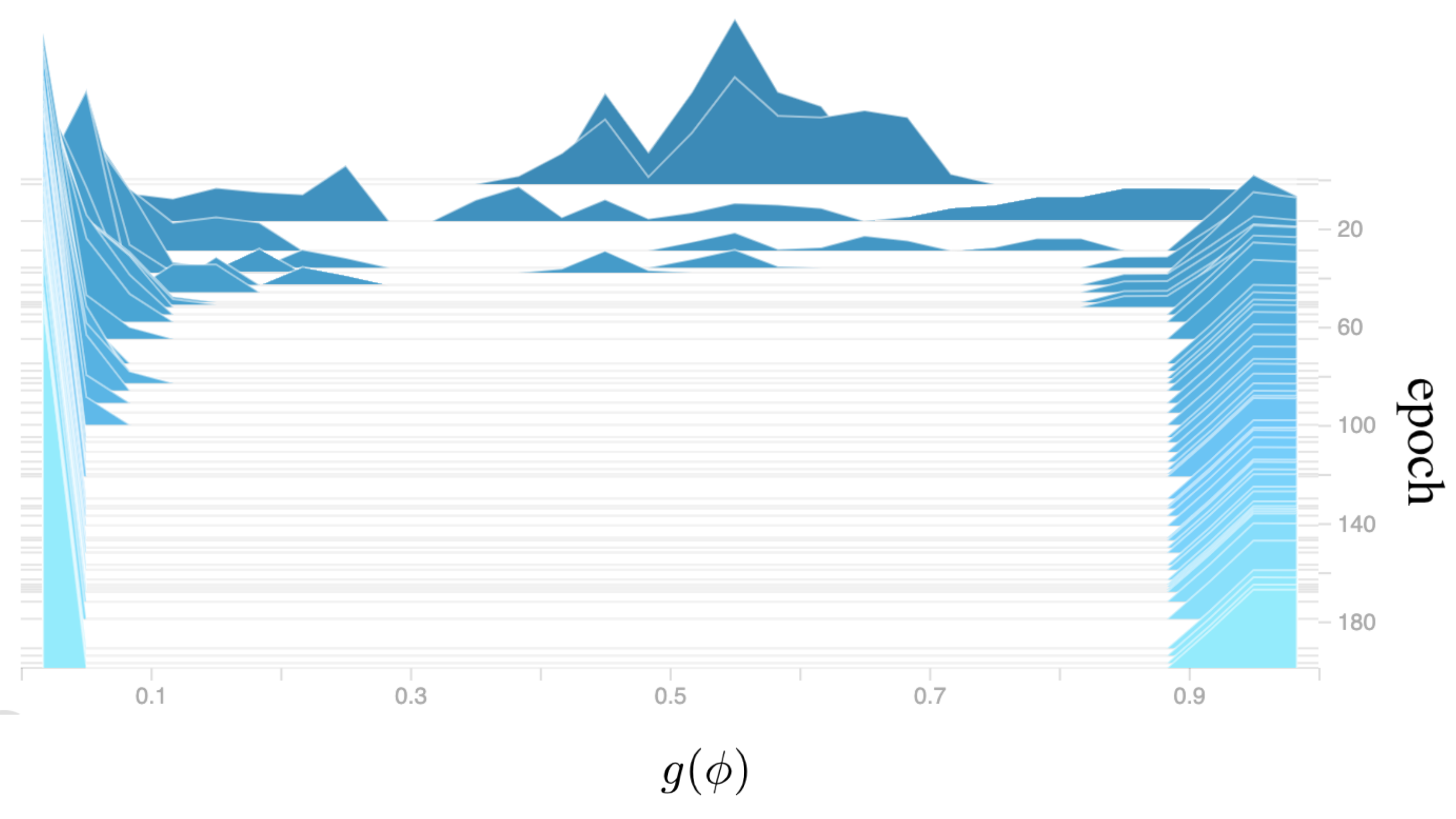}
\end{center}\vspace{-16pt}
\caption{Evolution of the histogram of $g(\phi)$ over training epochs. All $g(\phi)$ are initialized by random samples from a normal distribution $\mathcal{N}(0.5, 0.01)$, which are split into two spikes during training.}\label{fig:mountain}
\end{figure}

After training, we get model parameters $\tilde{\bs{\theta}}$ and $\bs{\phi}$. At test time, we can use the expectation of $\bs{z}\sim\mathrm{Ber}(\bs{z}; g(\bs{\phi}))$ as the mask $\hat{\bs{z}}$ for the final model parameters $\hat{\bs{\theta}}$:
\begin{equation}\label{eq:Ez}
\hat{\bs{z}} = \mathbb{E}[\bs{z}] = g (\bs{\phi}), \quad\quad \hat{\bs{\theta}}=\bs{\tilde{\theta}}\odot\hat{\bs{z}}.
\end{equation} 
However, this will not yield a sparsified network for inference since none of the element of $\hat{\bs{z}}=g(\bs{\phi})$ is exact zero (unless the hard sigmoid gate $g_{\bar{\sigma}_k}(\phi)$ is used). A simple approximation is to set the elements of $\hat{\bs{z}}$ to zero if the corresponding values in $g(\bs{\phi})$ are less than a threshold $\tau$, i.e.,
\begin{align}\label{eq:zbar}
  \bar{z}_j = \left\{
  \begin{array}{lr}
     0, & g(\phi_j) \le \tau   \\
     g(\phi_j), &\text{otherwise\;} 
  \end{array}
  \quad\quad j=1,2,\cdots,|\bs {z}| \right.
\end{align}
We find that this approximation is very effective in all of our experiments as the histogram of $g(\bs{\phi})$ is widely split into two spikes around values of 0 and 1 after training because of the sharp transition of the scaled sigmoid (or hard sigmoid) function. See Fig.~\ref{fig:mountain} for a typical plot of the histograms of $g(\bs{\phi})$ evolving during training process. We notice that our algorithm isn't very sensitive to $\tau$, tuning which incurs negligible impacts to prune rates and model accuracies. Therefore, for all of our experiments we set $\tau=0.5$ by default. Apparently, better designed $\tau$ is possible by considering the histogram of $g(\bs{\phi})$. However, we find this isn't very necessary for all of our experiments in the paper. Therefore, we will consider this histogram-dependent $\tau$ as our future improvement.

\subsection{Imposing Shrinkage on Model Parameters $\bs{\theta}$}\label{sec:L2}
The $L_0$ regularized objective function~(\ref{eq:L0-obj}) leads to sparse estimate of model parameters without imposing any shrinkage on the magnitude of $\bs{\theta}$. In some cases it might still be desirable to regularize the magnitude of model parameters with other norms, such as  $L_1$ or $L_2$ (weight decay), to improve the robustness of model. This can be achieved conveniently by computing the expected $L_1$ or $L_2$ norm of $\bs{\theta}$ under the same Bernoulli distribution: $\bs{z}\sim\mathrm{Ber}(\bs{z};g(\bs{\phi}))$ as follows: 
\begin{align}
  \mathbb{E}_{\bs{z}\sim\mathrm{Ber}\left(\bs{z}; g\left(\bs{\phi}\right)\right)}\left[||\bs\theta||_1\right]&=\sum_{j=1}^{|\theta|}\mathbb{E}_{z_j\sim\mathrm{Ber}(z_j;g(\phi_j))}\left[z_j|\tilde{\theta}_j|\right]=\sum_{j=1}^{|\theta|}g(\phi_j)|\tilde{\theta}_j|, \\
  \mathbb{E}_{\bs{z}\sim\mathrm{Ber}\left(\bs{z}; g\left(\bs{\phi}\right)\right)}\left[||\bs\theta||_2^2\right]&=\sum_{j=1}^{|\theta|}\mathbb{E}_{z_j\sim\mathrm{Ber}(z_j;g(\phi_j))}\left[z_j^2\tilde{\theta}_j^2\right]=\sum_{j=1}^{|\theta|}g(\phi_j)\tilde{\theta}_j^2,
\end{align}
which can be incorporated to Eq.~\ref{eq:L0-obj} as additional regularization terms.

\subsection{Group Sparsity Under $L_0$ and $L_2$ Norms}\label{sec:group}
The formulation so far promotes a weight-level sparsity for network architectures. This sparsification strategy can compress model and reduce memory footprint of a network. However, it will usually not lead to effective speedups because weight-sparsified networks require sparse matrix multiplication and irregular memory access, which make it extremely challenging to effectively utilize the parallel computing resources of GPUs and CPUs. For the purpose of computational efficiency, it's usually preferable to perform group sparsity instead of weight-level sparsity. Similar to~\cite{WenWuWan16,NekMolAsh17,Louizos2017}, we can achieve this by sharing a stochastic binary gate $z$ among all the weights in a group. For example, a group can be all fan-out weights of a neuron in fully connected layers or all weights of a convolution filter. With this, the group regularized $L_0$ and $L_2$ norms can be conveniently expressed as
\begin{align}\label{eq:group}
  \mathbb{E}_{\bs{z}\sim\mathrm{Ber}\left(\bs{z}; g\left(\bs{\phi}\right)\right)}\left[||\bs\theta||_0\right]&=\sum_{\text{g}=1}^{|G|}|\text{g}|g(\phi_{\text{g}})\\
      \mathbb{E}_{\bs{z}\sim\mathrm{Ber}\left(\bs{z}; g\left(\bs{\phi}\right)\right)}\left[||\bs\theta||_2^2\right]&=\sum_{\text{g}=1}^{|G|}\left(g(\phi_{\text{g}})\sum_{j=1}^{|\text{g}|}\tilde{\theta}_j^2\right)
\end{align}
where $|G|$ denotes the number of groups and $|\text{g}|$ denotes the number of weights of
group $\text{g}$. For the reason of computational efficiency, we perform this group sparsity in all of our experiments.

\section{Related Work}\label{sec:related}

It is well-known that DNNs are extremely compute and memory intensive. Recently, there has been an increasing interest to network sparsification~\cite{han2015learning,han2015deep,WenWuWan16,li2016pruning,louizos2017bayesian,molchanov2017variational,NekMolAsh17,Louizos2017} as the applications of DNNs to practical real-time systems, such as the IoT devices, call for compute and memory efficient networks. One of the earliest sparsification methods is to prune the redundant weights based on the magnitudes~\cite{lecun1990optimal}, which is proved to be effective in modern CNN~\cite{han2015learning}. Although weight sparsification is able to compress networks, it can barely improve computational efficiency due to unstructured sparsity~\cite{WenWuWan16}. Therefore, magnitude-based group sparsity is proposed~\cite{WenWuWan16,li2016pruning}, which can compress networks while reducing computation cost significantly. These magnitude-based methods usually proceed in three stages: pre-train a full network, prune the redundant weights or filters, and fine-tune the pruned model. As a comparison, our method $L_0$-ARM trains a sparsified network from scratch without pre-training and fine-tuning, and therefore is more preferable. 

Another category of sparsification methods is based on Bayesian statistics and information theory~\cite{molchanov2017variational,NekMolAsh17,louizos2017bayesian}. For example, inspired by variational dropout~\cite{kingma2015variational}, Molchanov et al. propose a method that unbinds the dropout rate, and also leads to sparsified networks~\cite{molchanov2017variational}. 

Recently, Louizos et al.~\cite{Louizos2017} propose to sparsify networks with $L_0$-norm. Since the $L_0$ regularization explicitly penalizes number of non-zero parameters, this method is conceptually very appealing. However, the non-differentiability of $L_0$ norm prevents an effective gradient-based optimization. Therefore, Louizos et al.~\cite{Louizos2017} propose a hard concrete gradient estimator for this optimization problem. Our work is built on top of their $L_0$ formulation. However, instead of using a hard concrete estimator, we investigate the Augment-Reinforce-Merge (ARM)~\cite{Yin2019}, a recently proposed unbiased estimator, to this binary optimization problem.

\section{Experimental Results}\label{sec:exp}
We evaluate the performance of $L_0$-ARM and $L_0$-AR on multiple public datasets and multiple network architectures. Specifically, we evaluate MLP 500-300~\cite{lecun1998gradient} and LeNet 5-Caffe~\footnote{\url{https://github.com/BVLC/caffe/tree/master/examples/mnist}} on the MNIST dataset~\cite{mnist}, and Wide Residual Networks~\cite{zagoruyko2016wide} on the CIFAR-10 and CIFAR-100 datasets~\cite{cifar10}. For baselines, we refer to the following state-of-the-art sparsification algorithms: Sparse Variational Dropout (Sparse VD)~\cite{molchanov2017variational}, Bayesian Compression with group normal-Jeffreys (BC-GNJ) and group horseshoe (BC-GHS)~\cite{louizos2017bayesian}, and $L_0$-norm regularization with hard concrete estimator ($L_0$-HC)~\cite{Louizos2017}. For a fair comparison, we closely follow the experimental setups of $L_0$-HC~\footnote{\url{https://github.com/AMLab-Amsterdam/L0_regularization}}.

\subsection{Implementation Details}
We incorporate $L_0$-ARM and $L_0$-AR into the architectures of MLP, LeNet-5 and Wide ResNet. As we described in Sec.~\ref{sec:group}, instead of sparsifying weights, we apply group sparsity on neurons in fully-connected layers or on convolution filters in convolutional layers. Once a neuron or filter is pruned, all related weights are removed from the networks.  

The Multi-Layer Perceptron (MLP)~\cite{lecun1998gradient} has two hidden layers of size 300 and 100, respectively. We initialize $g(\bs{\phi})=\bs{\pi}$  by random samples from a normal distribution $\mathcal{N}(0.8, 0.01)$ for the input layer and $\mathcal{N}(0.5, 0.01)$ for the hidden layers, which activate around 80\% of neurons in input layer and around 50\% of neurons in hidden layers. LeNet-5-Caffe consists of two convolutional layers of 20 and 50 filters interspersed with max pooling layers, followed by two fully-connected layers with 500 and 10 neurons. We initialize $g(\bs{\phi})=\bs{\pi}$ for all neurons and filters by random samples from a normal distribution $\mathcal{N}(0.5, 0.01)$. Wide-ResNets (WRNs)~\cite{zagoruyko2016wide} have shown state-of-the-art performance on many image classification benchmarks. Following \cite{Louizos2017}, we only apply $L_0$ regularization on the first convolutional layer of each residual block, which allows us to incorporate $L_0$ regularization without further modifying residual block architecture. The architectural details of WRN are listed in Table~\ref{WRNArch}. For initialization, we activate around 70\% of convolutional filters.

\begin{table}[h]
  \caption{Architectural details of WRN incorporated with $L_{0}$-ARM. The number in parenthesis is the size of activation map of each layer. For brevity, only the modified layers are included.}\label{WRNArch}
  \centering
  \begin{tabular}{ll}
  \toprule
  Group name & Layers \\
  \midrule
  conv1 & [Original Conv (16)] \\
  conv2 & [$L_0$ ARM (160); Original Conv (160)] $\times$ 4 \\
  conv3 & [$L_0$ ARM (320); Original Conv (320)] $\times$ 4 \\
  conv4 & [$L_0$ ARM (640); Original Conv (640)] $\times$ 4 \\
  \bottomrule
  \end{tabular}
\end{table}

For MLP and LeNet-5, we train with a mini-batch of 100 data samples and use Adam~\cite{kingma2014adam} as optimizer with initial learning rate of $0.001$, which is halved every 100 epochs. For Wide-ResNet, we train with a mini-batch of 128 data samples and use Nesterov Momentum as optimizer with initial learning rate of $0.1$, which is decayed by 0.2 at epoch 60 and 120. Each of these experiments run for 200 epochs in total. For a fair comparison, these experimental setups closely follow what were described in $L_0$-HC~\cite{Louizos2017} and their open-source implementation $^3$.

\subsection{MNIST Experiments}

We run both MLP and LeNet-5 on the MNIST dataset. By tuning the regularization strength $\lambda$, we can control the trade off between sparsity and accuracy. We can use one $\lambda$ for all layers or a separate $\lambda$ for each layer to fine-tune the sparsity preference. In our experiments, we set $\lambda=0.1/N$ or $\lambda=(0.1, 0.3, 0.4)/N$ for MLP, and set $\lambda=0.1/N$ or $\lambda=(10, 0.5, 0.1, 10)/N$ for LeNet-5, where $N$ denotes to the number of training datapoints.

We use three metrics to evaluate the performance of an algorithm: prediction accuracy, prune rate, and expected number of floating point operations (FLOPs). Prune rate is defined as the ratio of number of pruned weights to number of all weights. Prune rate manifests the memory saving of a sparsified network, while expected FLOPs demonstrates the training / inference cost of a sparsification algorithm.

\begin{table}[h]
  \caption{Performance comparison on MNIST. Each experiment was run five times and the median (in terms of accuracy) is reported. All the baseline results are taken from the corresponding papers.}
  \label{MLP&LENET}
  \centering
  \begin{tabular}{clccc}
  \toprule
  Network & Method & Pruned Architecture & Prune rate (\%) & Accuracy (\%) \\
  \midrule
  \multirow{4}*{\tabincell{cc}{MLP\\ 784-300-100}}  & Sparse VD & 219-214-100 & 74.72 & 98.2\\
  ~ & BC-GNJ & 278-98-13 & 89.24 & 98.2\\
  ~ & BC-GHS & 311-86-14 & 89.45 & 98.2\\
  ~ & $L_0$-HC ($\lambda = 0.1/N$) & 219-214-100 & 73.98 & 98.6\\
  ~ & $L_0$-HC ($\lambda$ sep.) & 266-88-33 & 89.99 & 98.2\\
  \cline{2-5}
  ~ & $L_0$-AR ($\lambda = 0.1/N$) & 453-150-68 & 70.39 & 98.3 \\
  ~ & $L_0$-ARM ($\lambda = 0.1/N$) & 143-153-78 & 87.00 & 98.3 \\
  ~ & $L_0$-AR ($\lambda$ sep.) & 464-114-65 & 77.10 & 98.2\\
  ~ & $L_0$-ARM ($\lambda$ sep.) & 159-74-73 & \textbf{92.96} & 98.1 \\
  \midrule
  \multirow{4}*{\tabincell{cc}{LeNet-5-Caffe\\ 20-50-800-500}} & Sparse VD & 14-19-242-131 & 90.7 & 99.0\\
  ~ & GL & 3-12-192-500 & 76.3 & 99.0\\
  ~ & GD & 7-13-208-16 & 98.62 & 99.0\\
  ~ & SBP & 3-18-284-283 & 80.34 & 99.0\\
  ~ & BC-GNJ & 8-13-88-13 & 99.05 & 99.0\\
  ~ & BC-GHS & 5-10-76-16 & 99.36 & 99.0\\
  ~ & $L_0$-HC ($\lambda = 0.1/N$) & 20-25-45-462 & 91.1 & 99.1\\      
  ~ & $L_0$-HC ($\lambda$ sep.) & 9-18-65-25 & 98.6 & 99.0\\

  \cline{2-5}
  ~ & $L_0$-AR ($\lambda = 0.1/N$) & 18-28-46-249 & 93.73 & 98.8\\  
  ~ & $L_0$-ARM ($\lambda = 0.1/N$) & 20-16-32-257 & 95.52 & 99.1\\  
  ~ & $L_0$-AR ($\lambda$ sep.) & 5-12-131-22 & 98.90 & 98.4\\  
  ~ & $L_0$-ARM ($\lambda$ sep.) & 6-10-39-11 & \textbf{99.49} & 98.7\\    
  \bottomrule
  \end{tabular}
\end{table}

We compare $L_0$-ARM and $L_0$-AR to five state-of-the-art sparsification algorithms on MNIST, with the results shown in Table~\ref{MLP&LENET}. For the comparison between $L_0$-HC and $L_0$-AR(M) when $\lambda=0.1/N$, we use the exact same hyper-parameters for both algorithms (the fairest comparison). In this case, $L_0$-ARM achieve the same accuracy (99.1\%) on LeNet-5 with even sparser pruned architectures (95.52\% vs. 91.1\%). When separated $\lambda$s are considered ($\lambda$ sep.), since $L_0$-HC doesn't disclose the specific $\lambda$s for the last two fully-connected layers, we tune them by ourselves and find that $\lambda=(10, 0.5, 0.1, 10)/N$ yields the best performance. In this case, $L_0$-ARM achieves the highest prune rate (99.49\% vs. 98.6\%) with very similar accuracies (98.7\% vs. 99.1\%) on LeNet-5. Similar patterns are also observed on MLP. Regarding $L_0$-AR, although its performance is not as good as $L_0$-ARM, it's still very competitive to all the other methods. The advantage of $L_0$-AR over $L_0$-ARM is its lower computational complexity during training. As we discussed in Sec.~\ref{sec:algo}, $L_0$-ARM needs one extra forward pass to estimate the gradient w.r.t. $\bs{\phi}$; for large DNN architectures, this extra cost can be significant. 

\begin{figure}[h]
\begin{center}
  \subfigure[$\lambda=0.1/N$]{\includegraphics[width=0.48\linewidth]{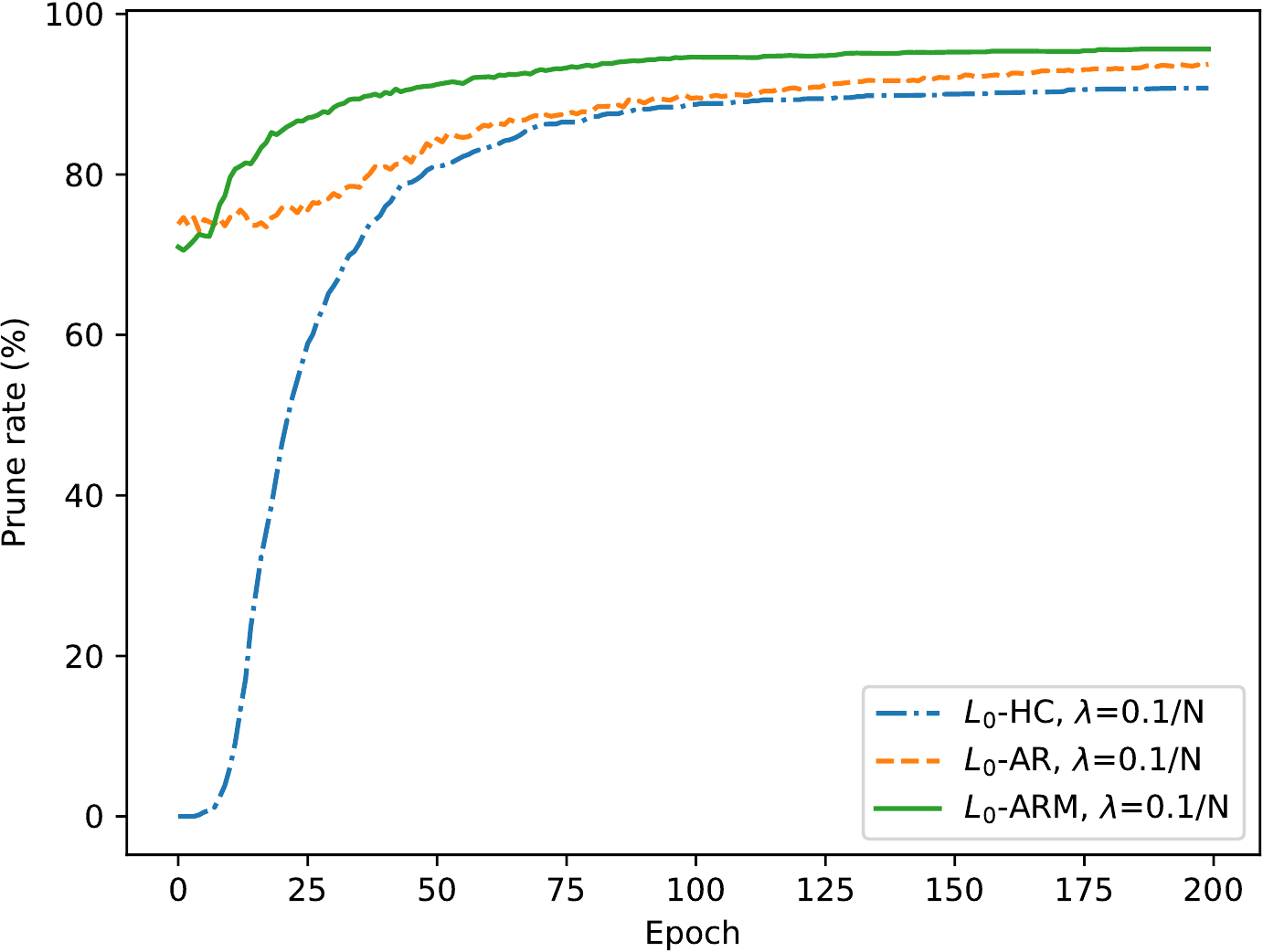}}\hfill
  \subfigure[$\lambda= sep.$]{\includegraphics[width=0.48\linewidth]{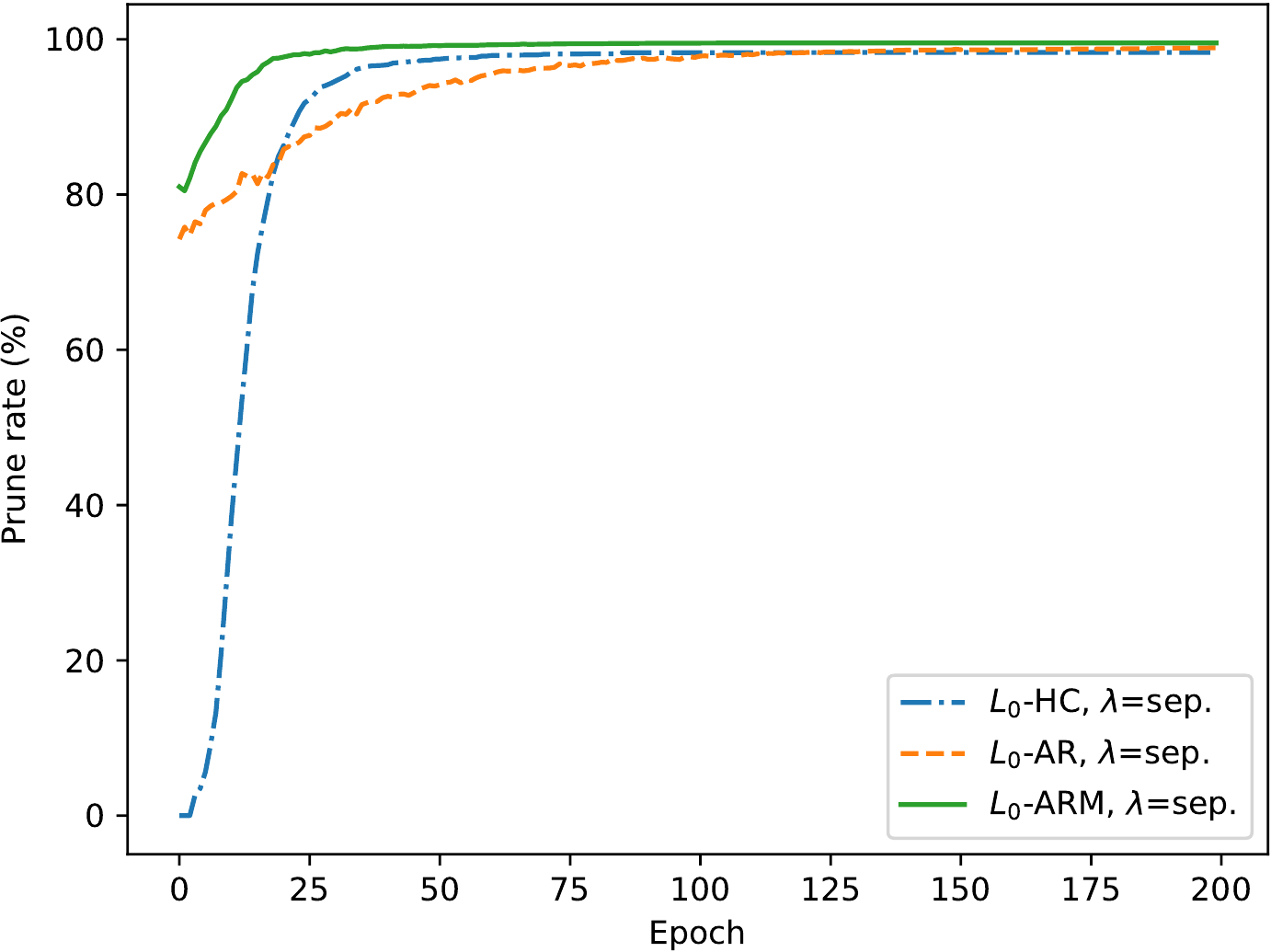}}
      \subfigure[$\lambda=0.1/N$]{\includegraphics[width=0.48\linewidth]{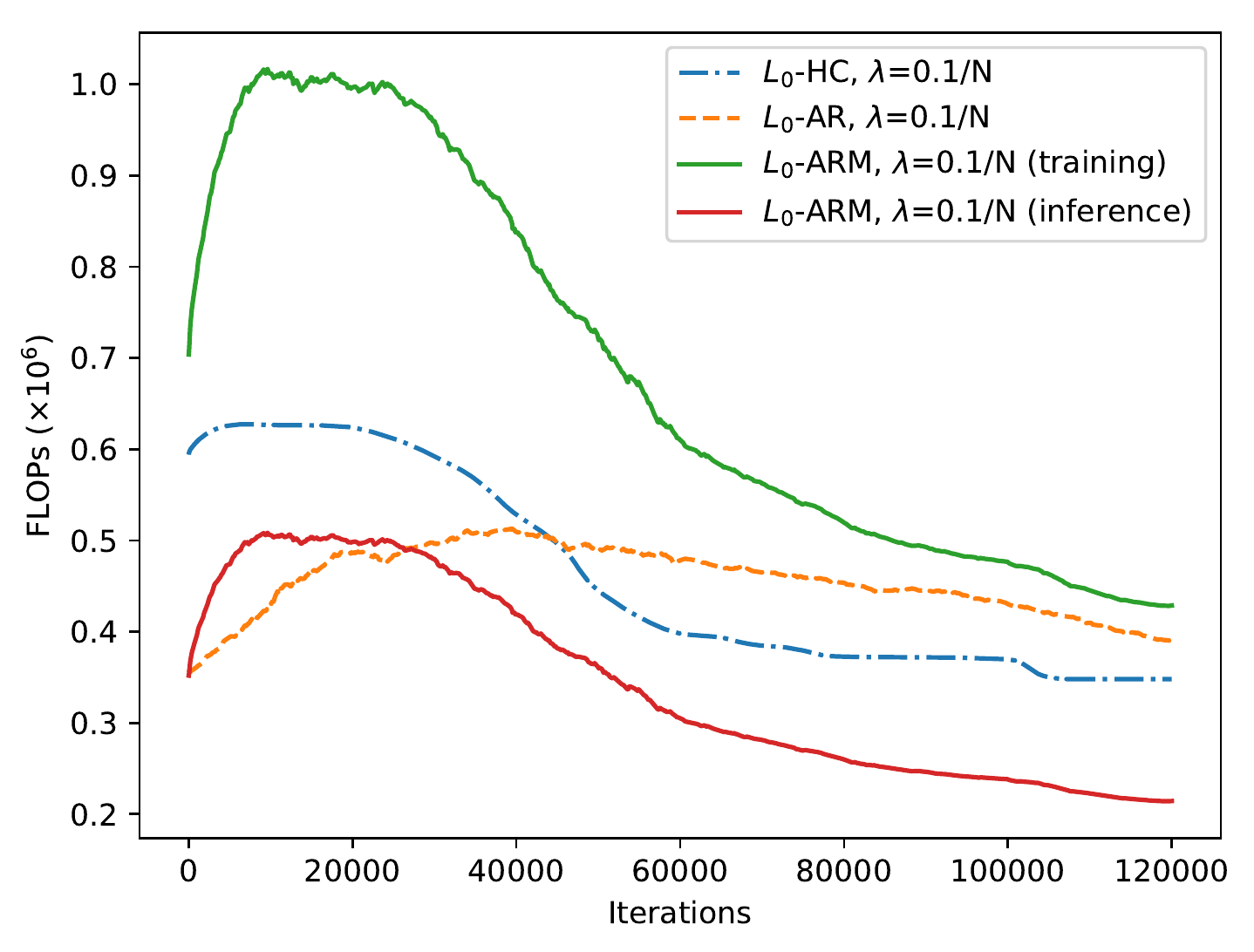}}
  \hfill
  \subfigure[$\lambda= sep.$]{\includegraphics[width=0.48\linewidth]{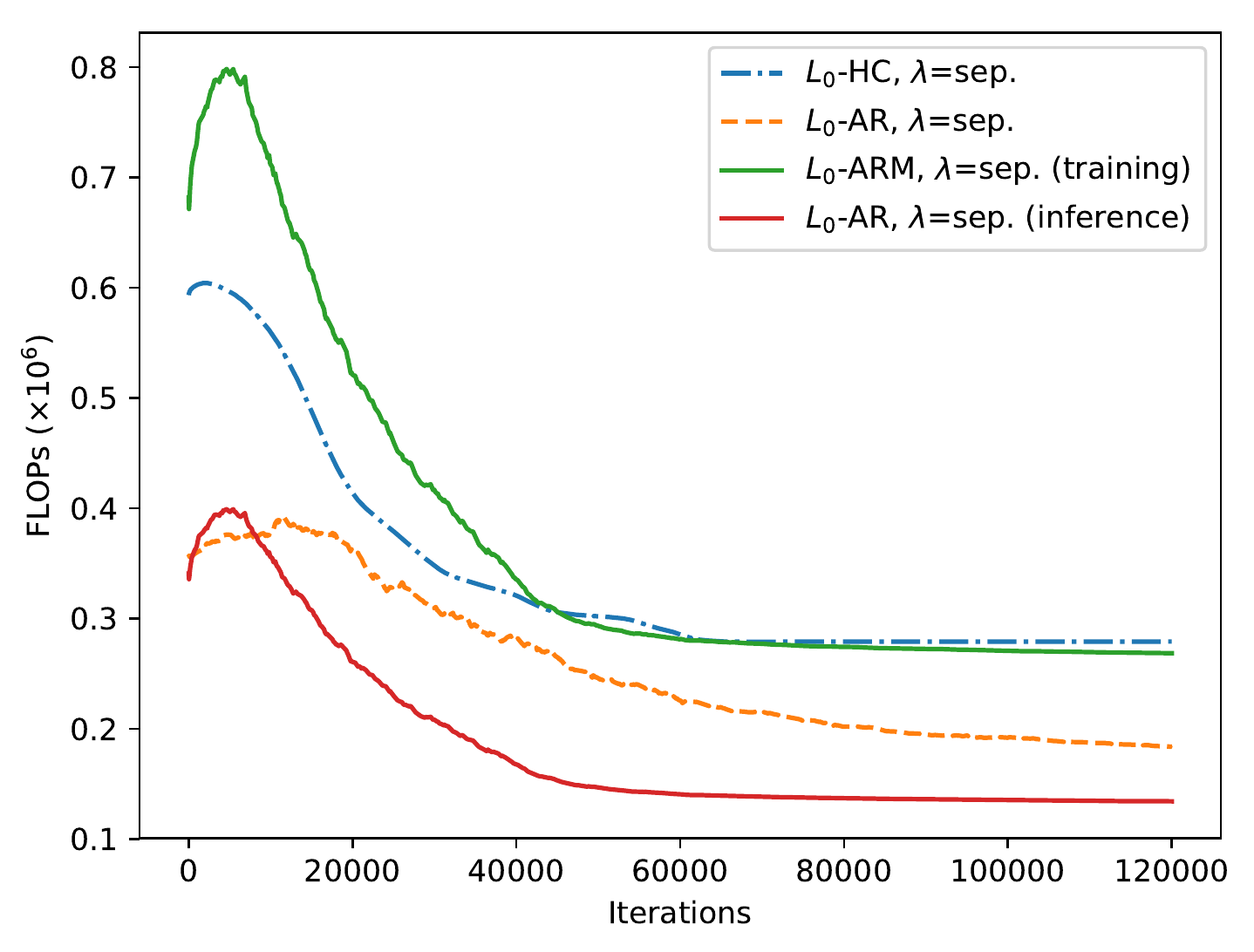}}
\end{center}\vspace{-16pt}
\caption{(a, b) Comparison of prune rate of sparsified network as a function of epoch for different algorithms. (c, d) Comparison of expected FLOPs as a function of epoch for different algorithms during training and inference. The results are on LeNet-5 with $L_0$-HC and $L_0$-AR(M). Because $L_0$-HC and $L_0$-AR only need one forward pass to compute gradient, they have the same expected FLOPs for training and inference. $L_0$-ARM needs two forward passes for training. Therefore, $L_0$-ARM is computationally more expensive during training (red curves), but it leads to sparser / more efficient architectures for inference (blue curves), which pays off its extra cost in training.}\label{fig:rate&flops}
\end{figure}

To evaluate the training cost and network sparsity of different algorithms, we compare the prune rates of $L_0$-HC and $L_0$-AR(M) on LeNet-5 as a function of epoch in Fig.~\ref{fig:rate&flops}~(a, b). Similarly, we compare the expected FLOPs of different algorithms as a function of epoch in Fig.~\ref{fig:rate&flops}~(c, d). As we can see from (a, b), $L_0$-ARM yields much sparser network architectures over the whole training epochs, followed by $L_0$-AR and $L_0$-HC. The FLOPs vs. Epoch plots in (c, d) are more complicated. Because $L_0$-HC and $L_0$-AR only need one forward pass to compute gradient, they have the same expected FLOPs for training and inference. $L_0$-ARM needs two forward passes for training. Therefore, $L_0$-ARM is computationally more expensive during training (red curves), but it leads to sparser / more efficient architectures for inference (green curves), which pays off its extra cost in training.

\subsection{CIFAR Experiments}
We further evaluate the performance of $L_0$-ARM and $L_0$-AR with Wide-ResNet~\cite{zagoruyko2016wide} on CIFAR-10 and CIFAR-100. Following \cite{Louizos2017}, we only apply $L_0$ regularization on the first convolutional layer of each residual block, which allows us to incorporate $L_0$ regularization without further modifying residual block architecture. 

Table~\ref{CIFAR} shows the performance comparison between $L_0$-AR(M) and three baseline methods. We find that $L_0$-HC cannot sparsify the Wide-ResNet architecture (prune rate 0\%)~\footnote{This was also reported recently in the appendix of ~\cite{GalElsHoo19}, and can be easily reproduced by using the open-source implementation of $L_0$-HC $^3$.}, while $L_0$-ARM and $L_0$-AR prune around 50\% of the parameters of the impacted subnet. As we activate 70\% convolution filters in initialization, the around 50\% prune rate is not due to initialization. We also inspect the histograms of $g(\bs{\phi})$: As expected, they are all split into two spikes around the values of 0 and 1, similar to the histograms shown in Fig.~\ref{fig:mountain}. In terms of accuracies, both $L_0$-ARM and $L_0$-AR achieve very similar accuracies as the baseline methods.

\begin{table}[h]
  \caption{Performance comparison of WRN on CIFAR-10 and CIFAR-100. Each experiment was run five times and the median (in terms of accuracy) is reported. All the baseline results are taken from the corresponding papers. Only the architectures of pruned layers are shown.}\label{CIFAR}
  \centering
  \begin{tabular}{clccc}
  \toprule
  Network & Method & Pruned Architecture & Prune rate (\%) & Accuracy (\%) \\
  \hline
  \multirow{4}*{\tabincell{cc}{WRN-28-10\\CIFAR-10}} & Original WRN \cite{zagoruyko2016wide} & full model & 0 & 96.00\\
  ~ & Original WRN-dropout \cite{zagoruyko2016wide} & full model & 0 & 96.11\\
  ~ & $L_0$-HC ($\lambda = 0.001/N$) \cite{Louizos2017} & full model & 0 & 96.17\\
  ~ & $L_0$-HC ($\lambda = 0.002/N$) \cite{Louizos2017} & full model & 0 & 96.07\\
  \cline{2-5}
  ~ & $L_0$ AR ($\lambda = 0.001/N$) & \tabincell{cc}{83-77-83-88-\\169-167-153-165-\\324-323-314-329} & 49.49 & 95.58\\
  \cline{2-5}
  ~ & $L_0$ ARM ($\lambda = 0.001/N$) & \tabincell{cc}{74-86-83-83-\\164-145-167-153-\\333-333-310-330} & 49.46 & 95.68\\
  \cline{2-5}
  ~ & $L_0$ AR ($\lambda = 0.002/N$) & \tabincell{cc}
  {82-75-82-87-\\164-169-156-161-\\317-317-317-324} & \textbf{49.95} & 95.60\\
  \cline{2-5}
  ~ & $L_0$ ARM ($\lambda = 0.002/N$) & \tabincell{cc}{75-72-78-78-\\157-165-131-162-\\336-325-331-343} & 49.63 & 95.70\\
  \hline
  \multirow{4}*{\tabincell{cc}{WRN-28-10\\CIFAR-100}} & Original WRN \cite{zagoruyko2016wide}  & full model & 0 & 78.82 \\
  ~ & Original WRN-dropout \cite{zagoruyko2016wide}  & full model & 0 & 81.15 \\
  ~ & $L_0$-HC ($\lambda = 0.001/N$) \cite{Louizos2017}  & full model & 0 & 81.25\\
  ~ & $L_0$-HC ($\lambda = 0.002/N$) \cite{Louizos2017}  & full model & 0 & 80.96\\
  \cline{2-5}
  ~ & $L_0$-AR ($\lambda = 0.001/N$) & \tabincell{cc}
  {78-78-79-85-\\168-168-162-164-\\308-326-319-330} & 49.37 & 80.50\\
  \cline{2-5}
  ~ & $L_0$-ARM ($\lambda = 0.001/N$) & \tabincell{cc}{75-83-80-58-\\172-156-160-165-\\324-311-313-318} & 50.51 & 80.74 \\
  \cline{2-5}
  ~ & $L_0$-AR ($\lambda = 0.002/N$) & \tabincell{cc}{75-76-72-80-\\158-158-137-168-\\318-295-327-324} & 50.93 & 80.09 \\
  \cline{2-5}
  ~ & $L_0$-ARM ($\lambda = 0.002/N$) & \tabincell{cc}{81-74-77-73-\\149-157-156-152-\\299-332-305-325} & \textbf{50.78} & 80.56\\
  \bottomrule
  \end{tabular}
\end{table}

\begin{figure}[h]
  \begin{center}
  \subfigure[CIFAR-10]{\includegraphics[width=0.48\linewidth]{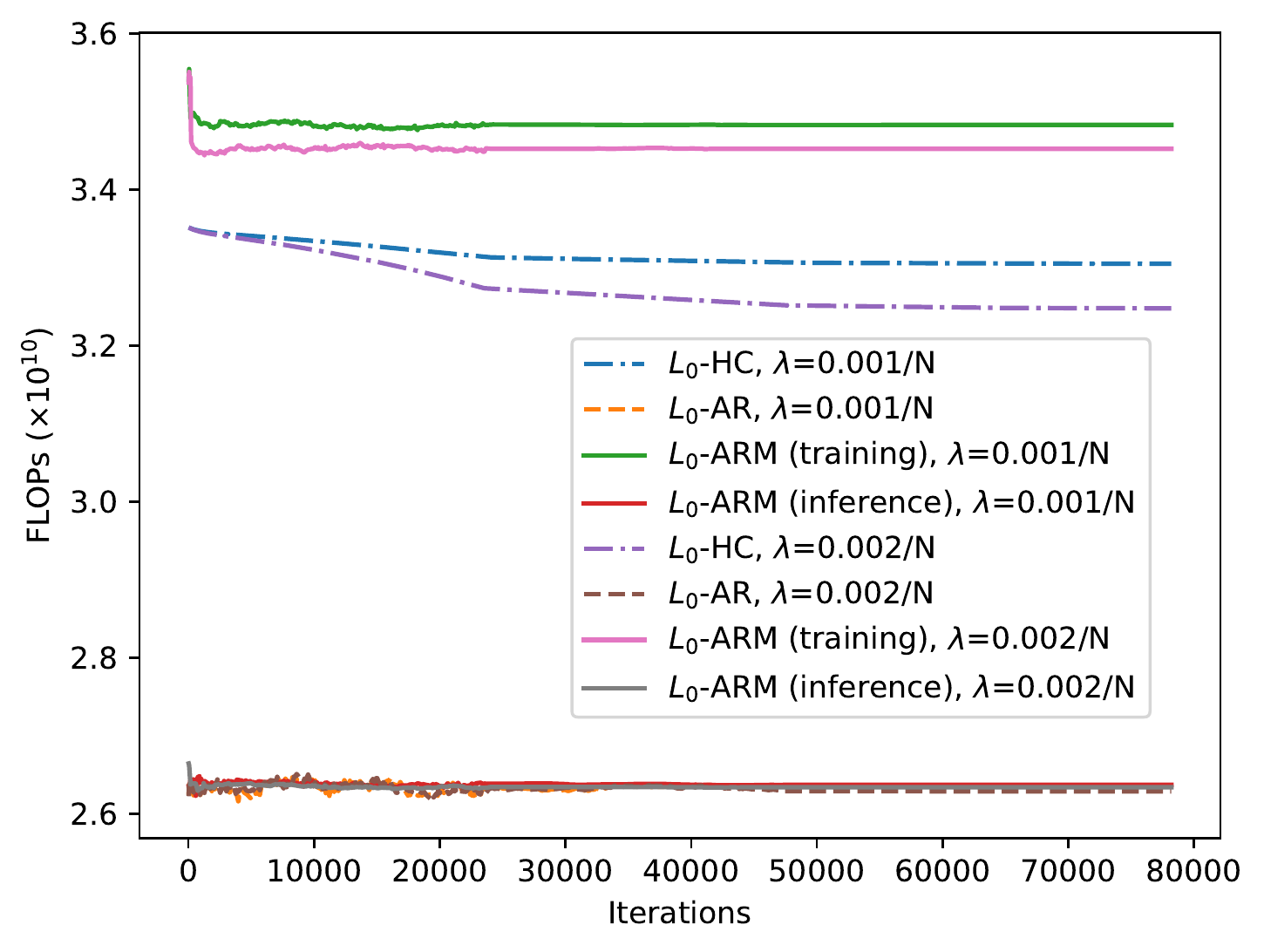}}
  \hfill
  \subfigure[CIFAR-100]{\includegraphics[width=0.48\linewidth]{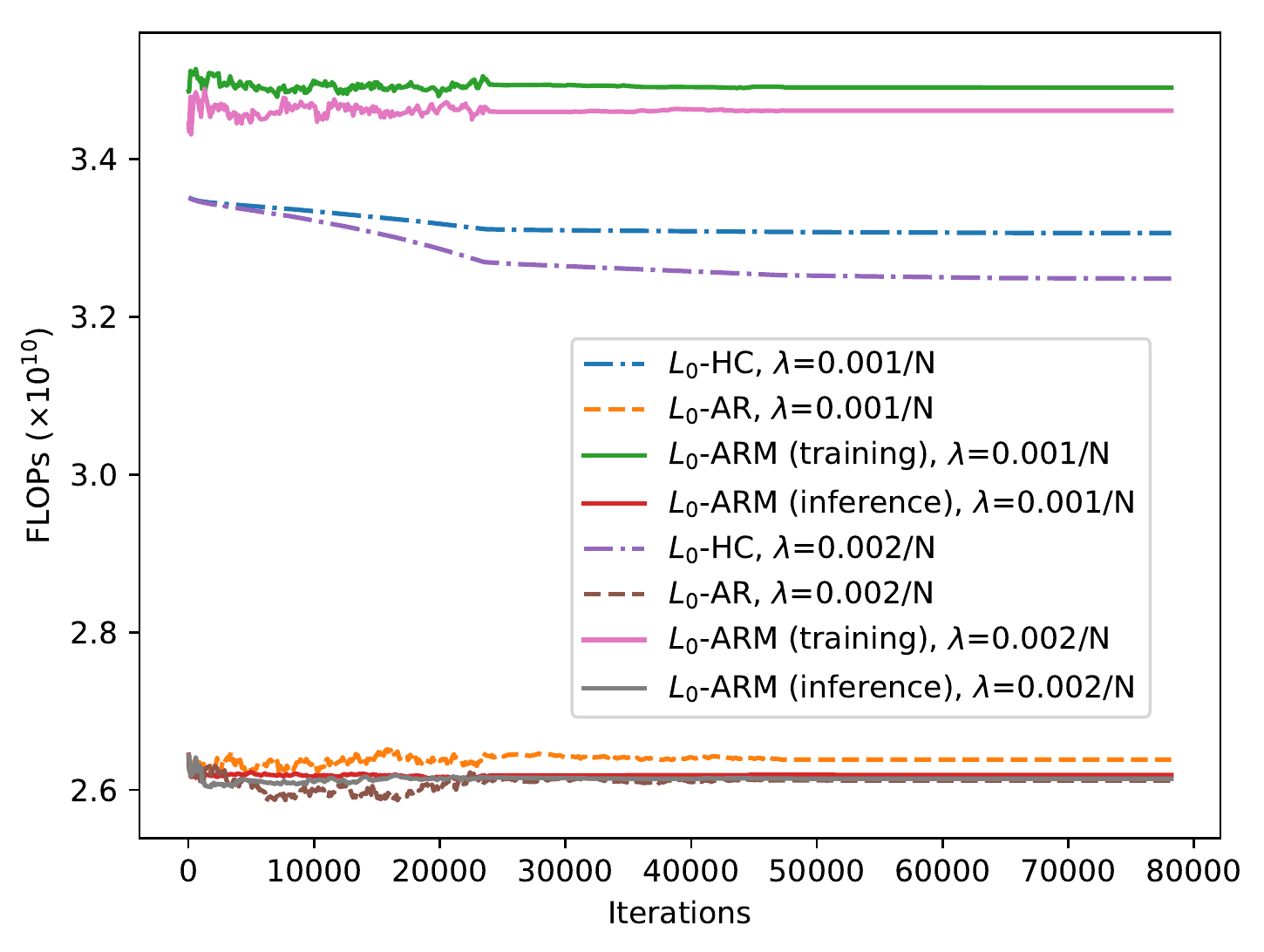}}
  \end{center}\vspace{-16pt}
  \caption{Comparison of expected FLOPs as a function of iteration during training and inference. Similar to Fig.~\ref{fig:rate&flops}, $L_0$-ARM is more computationally expensive for training, but leads to sparser / more efficient architectures for inference. For these experiments, $L_0$-AR has the lowest training FLOPs and inference FLOPs, while achieving very similar accuracies as the baseline methods (Table ~\ref{CIFAR}).}\label{fig:cifar_flops}
\end{figure}

To evaluate the training and inference costs of different algorithms, we compare the expected FLOPs of $L_0$-HC and $L_0$-AR(M) on CIFAR-10 and CIFAR-100 as a function of iteration in Fig.~\ref{fig:cifar_flops}. Similar to Fig.~\ref{fig:rate&flops}, $L_0$-ARM is more computationally expensive for training, but leads to sparser / more efficient architectures for inference, which pays off its extra cost in training. It's worth to emphasize that for these experiments $L_0$-AR has the lowest training FLOPs and inference FLOPs (since only one forward pass is needed for training and inference), while achieving very similar accuracies as the baseline methods (Table~\ref{CIFAR}).

\begin{figure}[h]
  \begin{center}
  \includegraphics[width=0.9\linewidth]{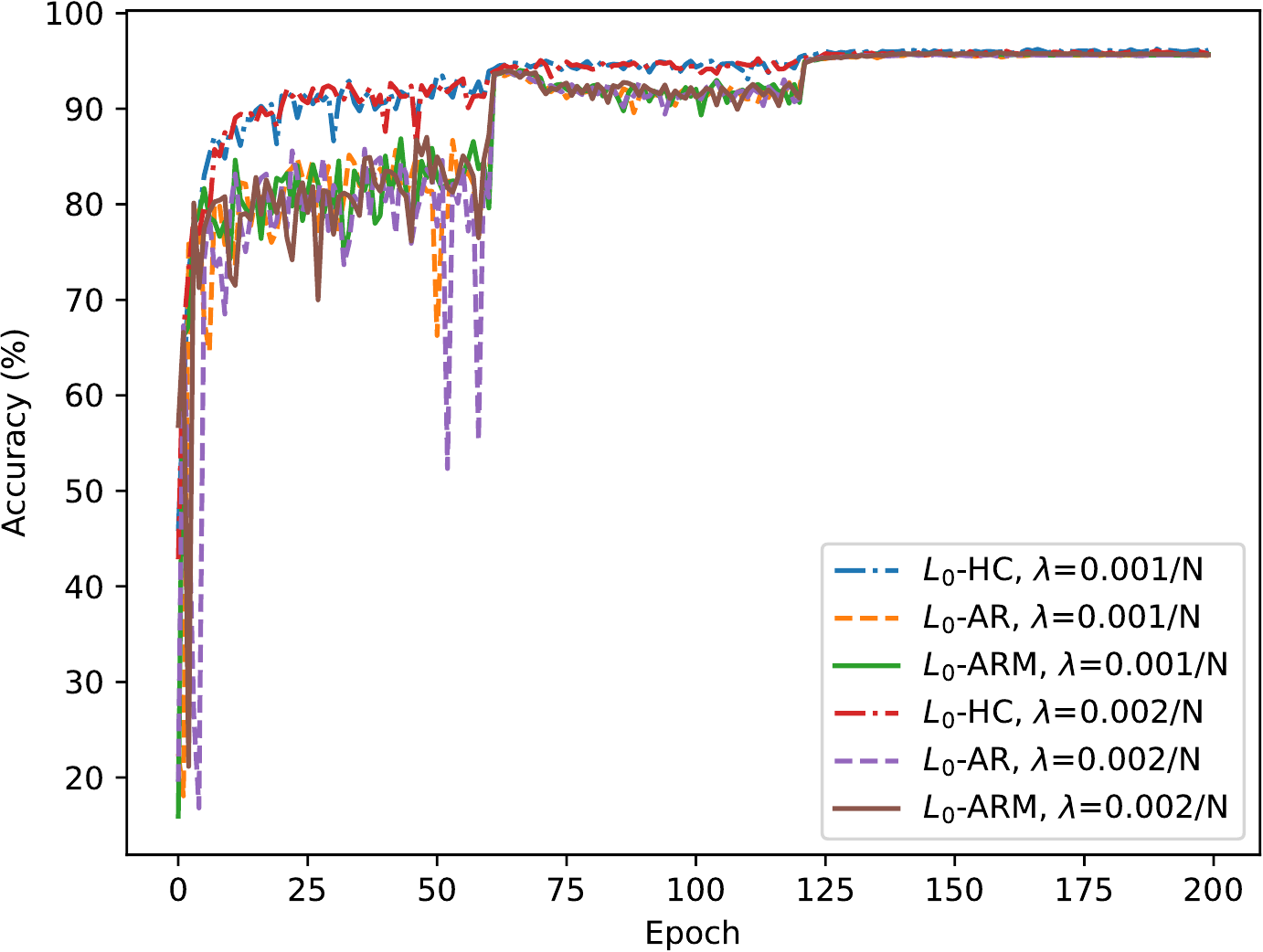}
  \end{center}\vspace{-16pt}
  \caption{Comparison of test accuracy as a function of epoch for different algorithms on CIFAR-10. We apply the exact same hyper-parameters of $L_0$-HC to $L_0$-AR(M), which yield similar accuracies for converged models even though the latter prunes around 50\% parameters while the former prunes 0\%.}\label{fig:cifar10_acc}
\end{figure}

Finally, we compare the test accuracies of different algorithms as a function of epoch on CIFAR-10, with the results shown in Fig.~\ref{fig:cifar10_acc}. We apply the exact same hyper-parameters of $L_0$-HC to $L_0$-AR(M). As $L_0$-AR(M) prunes around 50\% parameters during training (while $L_0$-HC prunes 0\%), the test accuracies of the former are lower than the latter before convergence, but all the algorithms yield very similar accuracies after convergence, demonstrating the effectiveness of $L_0$-AR(M).

\section{Conclusion}\label{sec:conclusion}
We propose $L_0$-ARM, an unbiased and low-variance gradient estimator, to sparsify network architectures. Compared to $L_0$-HC~\cite{Louizos2017} and other state-of-the-art sparsification algorithms, $L_0$-ARM demonstrates superior performance of sparsifying network architectures while retaining almost the same accuracies of the baseline methods. Extensive experiments on multiple public datasets and multiple network architectures validate the effectiveness of $L_0$-ARM. Overall, $L_0$-ARM yields the sparsest architectures and the lowest inference FLOPs for all the networks considered with very similar accuracies as the baseline methods.

As for future extensions, we plan to design better (possibly non-antithetic) parametric function $g(\phi)$ to improve the sparsity of solutions. We also plan to investigate more efficient algorithm to evaluate $L_0$-ARM gradient~(\ref{eq:L0-ARM}) by utilizing the antithetic structure of two forward passes. 

%
%
%
\bibliographystyle{splncs04}

\begin{thebibliography}{10}
  \providecommand{\url}[1]{\texttt{#1}}
  \providecommand{\urlprefix}{URL }
  \providecommand{\doi}[1]{https://doi.org/#1}
  
  \bibitem{BenLeoCou18}
  Bengio, Y., Leonard, N., Courville, A.: Estimating or propagating gradients
    through stochastic neurons for conditional computation. arXiv preprint
    arXiv:1308.3432  (2013)
  
  \bibitem{SVO18}
  Bird, T., Kunze, J., Barber, D.: Stochastic variational optimization. arXiv
    preprint arXiv:1809.04855  (2018)
  
  \bibitem{imagenet09}
  Deng, J., Dong, W., Socher, R., Li, L.J., Li, K., Fei-Fei, L.: Imagenet: A
    large-scale hierarchical image database. In: IEEE Conference on Computer
    Vision and Pattern Recognition (CVPR) (2009)
  
  \bibitem{bert18}
  Devlin, J., Chang, M.W., Lee, K., Toutanova, K.: Bert: Pre-training of deep
    bidirectional transformers for language understanding. arXiv preprint
    arXiv:1810.04805  (2018)
  
  \bibitem{GalElsHoo19}
  Gale, T., Elsen, E., Hooker, S.: The state of sparsity in deep neural networks.
    arXiv preprint arXiv:1902.09574  (2019)
  
  \bibitem{relax18}
  Grathwohl, W., Choi, D., Wu, Y., Roeder, G., Duvenaud, D.: Backpropagation
    through the void: Optimizing control variates for black-box gradient
    estimation. In: International Conference on Learning Representations (ICLR)
    (2018)
  
  \bibitem{han2015deep}
  Han, S., Mao, H., Dally, W.J.: Deep compression: Compressing deep neural
    networks with pruning, trained quantization and huffman coding. In:
    International Conference on Learning Representations (ICLR) (2016)
  
  \bibitem{han2015learning}
  Han, S., Pool, J., Tran, J., Dally, W.: Learning both weights and connections
    for efficient neural network. In: Advances in neural information processing
    systems. pp. 1135--1143 (2015)
  
  \bibitem{resnet16}
  He, K., Zhang, X., Ren, S., Sun, J.: Deep residual learning for image
    recognition. In: IEEE Conference on Computer Vision and Pattern Recognition
    (CVPR). pp. 770--778 (2016)
  
  \bibitem{densenet17}
  Huang, G., Liu, Z., van~der Maaten, L., Weinberger, K.Q.: Densely connected
    convolutional networks. In: IEEE Conference on Computer Vision and Pattern
    Recognition (CVPR) (2017)
  
  \bibitem{gumbel-softmax17}
  Jang, E., Gu, S., Poole, B.: Categorical reparameterization with
    gumbel-softmax. In: International Conference on Learning Representations
    (ICLR) (2017)
  
  \bibitem{kingma2014adam}
  Kingma, D.P., Ba, J.: Adam: A method for stochastic optimization. In:
    International Conference on Learning Representations (ICLR) (2015)
  
  \bibitem{kingma2015variational}
  Kingma, D.P., Salimans, T., Welling, M.: Variational dropout and the local
    reparameterization trick. In: Advances in Neural Information Processing
    Systems. pp. 2575--2583 (2015)
  
  \bibitem{cifar10}
  Krizhevsky, A.: Learning multiple layers of features from tiny images. Tech.
    rep. (2009)
  
  \bibitem{mnist}
  Lecun, Y., Bottou, L., Bengio, Y., Haffner, P.: Gradient-based learning applied
    to document recognition. In: Proceedings of the IEEE. pp. 2278--2324 (1998)
  
  \bibitem{lecun1998gradient}
  LeCun, Y., Bottou, L., Bengio, Y., Haffner, P., et~al.: Gradient-based learning
    applied to document recognition. Proceedings of the IEEE  \textbf{86}(11),
    2278--2324 (1998)
  
  \bibitem{lecun1990optimal}
  LeCun, Y., Denker, J.S., Solla, S.A.: Optimal brain damage. In: Advances in
    neural information processing systems. pp. 598--605 (1990)
  
  \bibitem{li2016pruning}
  Li, H., Kadav, A., Durdanovic, I., Samet, H., Graf, H.P.: Pruning filters for
    efficient convnets. arXiv preprint arXiv:1608.08710  (2016)
  
  \bibitem{louizos2017bayesian}
  Louizos, C., Ullrich, K., Welling, M.: Bayesian compression for deep learning.
    In: Advances in Neural Information Processing Systems. pp. 3288--3298 (2017)
  
  \bibitem{Louizos2017}
  Louizos, C., Welling, M., Kingma, D.P.: Learning sparse neural networks through
    $l_0$ regularization. In: International Conference on Learning
    Representations (ICLR) (2018)
  
  \bibitem{concrete17}
  Maddison, C.J., Mnih, A., Teh, Y.W.: The concrete distribution: A continuous
    relaxation of discrete random variables. In: International Conference on
    Learning Representations (ICLR) (2017)
  
  \bibitem{molchanov2017variational}
  Molchanov, D., Ashukha, A., Vetrov, D.: Variational dropout sparsifies deep
    neural networks. In: Proceedings of the 34th International Conference on
    Machine Learning-Volume 70. pp. 2498--2507. JMLR. org (2017)
  
  \bibitem{NekMolAsh17}
  Neklyudov, K., Molchanov, D., Ashukha, A., Vetrov, D.: Structured bayesian
    pruning via log-normal multiplicative noise. In: Advances in Neural
    Information Processing Systems (NIPS) (2017)
  
  \bibitem{alphago16}
  Silver, D., Huang, A., Maddison, C.J., Guez, A., Sifre, L., van~den Driessche,
    G., Schrittwieser, J., Antonoglou, I., Panneershelvam, V., Lanctot, M.,
    Dieleman, S., Grewe, D., Nham, J., Kalchbrenner, N., Sutskever, I.,
    Lillicrap, T., Leach, M., Kavukcuoglu, K., Graepel, T., Hassabis, D.:
    Mastering the game of go with deep neural networks and tree search. Nature
    \textbf{529},  484--503 (2016)
  
  \bibitem{rebar17}
  Tucker, G., Mnih, A., Maddison, C.J., Lawson, J., Sohl-Dickstein, J.: Rebar:
    Low-variance, unbiased gradient estimates for discrete latent variable
    models. In: Advances in Neural Information Processing Systems (NIPS) (2017)
  
  \bibitem{WenWuWan16}
  Wen, W., Wu, C., Wang, Y., Chen, Y., Li, H.: Learning structured sparsity in
    deep neural networks. In: Advances in Neural Information Processing Systems
    (NIPS) (2016)
  
  \bibitem{reinforce92}
  Williams, R.J.: Simple statistical gradient-following algorithms for
    connectionist reinforcement learning. Machine Learning  \textbf{8}(3-4),
    229--256 (May 1992)
  
  \bibitem{Yin2019}
  Yin, M., Zhou, M.: Arm: Augment-{REINFORCE}-merge gradient for stochastic
    binary networks. In: International Conference on Learning Representations
    (ICLR) (2019)
  
  \bibitem{zagoruyko2016wide}
  Zagoruyko, S., Komodakis, N.: Wide residual networks. In: The British Machine
    Vision Conference (BMVC) (2016)
  
  \end{thebibliography}

  %
\end{document}